\documentclass{article}
\pdfoutput=1

\usepackage[final]{nips_2016}

\usepackage[utf8]{inputenc} 
\usepackage[T1]{fontenc}    
\usepackage{url}            
\usepackage{booktabs}       
\usepackage{amsfonts}       
\usepackage{nicefrac}       
\usepackage{microtype}      

\usepackage{graphicx}
\usepackage{amsmath}
\usepackage{csquotes}
\usepackage{enumerate}
\usepackage{algorithm}
\usepackage[noend]{algpseudocode}
\usepackage{xcolor}
\usepackage{todonotes}

\DeclareMathOperator*{\argmax}{arg\, max}

\title{Safety-Aware Robot Damage Recovery \\Using Constrained Bayesian Optimization \\and Simulated Priors}

\author{
    Vaios Papaspyros\textsuperscript{1, 2, 3, 4} \\
    \And
    Konstantinos Chatzilygeroudis\textsuperscript{1, 2, 3} \\
    \And
    Vassilis Vassiliades\textsuperscript{1, 2, 3} \\
    \And
    Jean-Baptiste Mouret\textsuperscript{1, 2, 3,}\thanks{
    Corresponding author: jean-baptiste.mouret@inria.fr \newline
    \textsuperscript{1} Inria, Villers-lès-Nancy, F-54600, France.
    \textsuperscript{2} CNRS, Loria, UMR 7503, Vandœuvre-lès-Nancy, F-54500, France.
    \textsuperscript{3} Universitè de Lorraine, Loria, UMR 7503, Vandœuvre-lès-Nancy, F-54500, France.
    \textsuperscript{4} University of Patras, 26500 Rion, Patras, Greece
    }\\
}

\begin{document}

\maketitle

\begin{abstract}
  The recently introduced Intelligent Trial-and-Error (IT\&E) algorithm showed that robots can adapt to damage in a matter of a few trials. The success of this algorithm relies on two components: prior knowledge acquired through simulation with an intact robot, and Bayesian optimization (BO) that operates on-line, on the damaged robot. While IT\&E leads to fast damage recovery, it does not incorporate any safety constraints that prevent the robot from attempting harmful behaviors. In this work, we address this limitation by replacing the BO component with a constrained BO procedure. We evaluate our approach on a simulated damaged humanoid robot that needs to crawl as fast as possible, while performing as few unsafe trials as possible. We compare our new \enquote{safety-aware IT\&E} algorithm to IT\&E and a multi-objective version of IT\&E in which the safety constraints are dealt as separate objectives. Our results show that our algorithm outperforms the other approaches, both in crawling speed within the safe regions and number of unsafe trials.
\end{abstract}

\section{Introduction}
Current robots would greatly benefit from being capable of carrying on with their mission when they are damaged, as illustrated by the recent DARPA Robotics Challenge~\citep{atkeson_2016,carlson_how_2005}. When damaged, most robots attempt to diagnose the fault and search for a contingency plan; but they could also exploit a reinforcement learning algorithm
so that they can find
compensatory behaviors without having to "understand" the damage~\citep{cully2015robots,ren2015multiple}. If they follow this second approach, robots need learning algorithms that are highly data-efficient~\citep{chatzilygeroudis2016reset} because too many trials would waste precious time to achieve the mission.

One of the most promising approaches for data-efficient robot damage recovery is the recently introduced Intelligent Trial-and-Error algorithm (IT\&E)~\citep{cully2015robots}. It combines two ideas: (1) a Bayesian optimization (BO) algorithm~\citep{shahriari2016taking} that optimizes a reward function, because it is a generic, data-efficient policy search algorithm~\citep{calandra2014experimental}, and (2) a behavior-performance map generated before the mission with a simulation of the intact robot, which acts both as a prior for the Bayesian optimization algorithm and as a dimension reduction algorithm. This combination allowed a damaged 6-legged robot to find a new gait in about a dozen of trials (less than 2 minutes), and a robotic arm to overcome several blocked joints in a few minutes~\citep{cully2015robots}.

Unfortunately, trial-and-error approaches, like IT\&E, are likely to damage the robot even more because they will often try behaviors that are too extreme for the mechanical design. More generally, learning algorithms will push robots to their limits because they focus solely on maximizing the reward intake\footnote{Interestingly, learning algorithms also push robot simulators to their limits as they often exploit simulation inaccuracies.}. This issue is especially concerning for expensive prototypes like the iCub robot~\citep{metta2008icub, tsagarakis2007icub}: these robots are too expensive (around 250k euros for the iCub) and too fragile to try risky behaviors.

While recent methods, like SafeOPT~\citep{berkenkamp2016bayesian}, tackle this issue successfully, they require an initial safe set of parameters, that is hard to estimate in an unknown damage setting.
In this paper, we extend the IT\&E algorithm~\citep{cully2015robots} by adding safety constraints~\citep{gardner2014bayesian} and automatically computing priors over the safety of controller parameters, so that the probability of breaking the robot during the learning process is as low as possible. We evaluate our algorithm using a simulated damaged iCub robot.

\begin{figure}[t]
    \centering
    \includegraphics[width=0.92\textwidth]{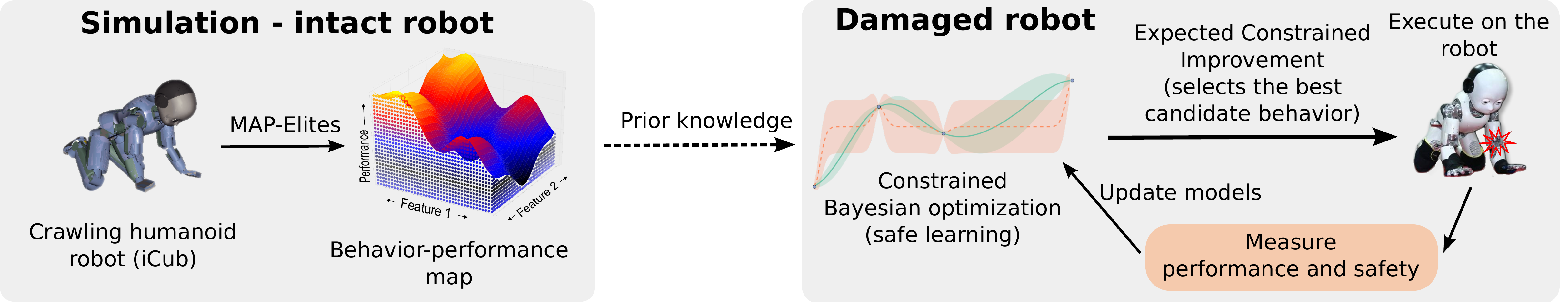}
    \caption{Overview of the safety-aware IT\&E algorithm. The algorithm first creates a behavior-performance map through MAP-Elites in simulation. This is fed as prior knowledge to a constrained BO procedure. The best candidate behavior is executed on the damaged robot while measuring the contact point forces and crawling speed. Finally, the Gaussian process models are updated.}
    \label{fig:concept}
\end{figure}

\section{Safety-aware Intelligent Trial \& Error Algorithm}


The first step of IT\&E is to create a low-dimensional behavior-performance map with a simulation of the intact robot. This step is achieved with an evolutionary algorithm called MAP-Elites~\citep{mouret2015illuminating}, which, instead of searching for a single, best solution, like standard optimization algorithms, searches for the highest-performing individual for each point in a user-defined space. This user-defined space is often called the behavior space, because the dimensions of variation (behavior descriptors) usually measure behavioral characteristics. For example, by defining one dimension for each leg's fraction of time spent on the ground, MAP-Elites produces a wide variety of walking gaits for a hexapod robot~\citep{cully2015robots}.

To search for the best behavior on the damaged robot, IT\&E alters the classical BO scheme~\citep{cully2015robots} by (1) searching a behavior in the map, instead of searching the best policy parameters, and (2) modeling the difference between the performance predicted by the map ($M(\cdot)$) and the actual performance, instead of directly modeling the performance function. Thus, at each step of the adaptation algorithm, the BO procedure selects the most promising behavior from the map, executes it on the damaged robot, observes the performance, and updates its predictions accordingly.

More precisely, the performance function, $f(\mathbf{x})$, is modeled as a Gaussian process (GP):
\begin{equation}
    f(\mathbf{x}) \sim GP(m(\mathbf{x}), k(\mathbf{x}, \mathbf{x'}))
\end{equation}
\begin{equation}
    p(f(\mathbf{x}) | X_{1:t},\mathbf{x}) = N(m(\mathbf{x}), \sigma^2(\mathbf{x}))
\end{equation}
where:
\begin{equation}
    m(\mathbf{x}) = M(\mathbf{x}) + \mathbf{k}^T(\mathbf{K} + \sigma_n^2 I)^{-1}(X_{1:t} - M(\mathbf{x}_{1:t}))
\end{equation}
\begin{equation}
    \sigma^2({\mathbf{x}}) = k({\mathbf{x}}, {\mathbf{x}}) - \mathbf{k}^T(\mathbf{K} + \sigma_n^2 I)^{-1}\mathbf{k}
\end{equation}
and $\mathbf{K}$ is the kernel matrix with $K_{ij} = k(\mathbf{x}_i, \mathbf{x}_j)$, $\mathbf{k} = k(X_{1:t}, \mathbf{x})$ and $X_{1:t}$ the set of observations.

This approach leads to short adaptation times in several experiments with damaged robots~\citep{cully2015robots}, but it has a serious limitation that prevents it from being used with expensive robots: it lacks safety constraints, that is, nothing prevents the robot from trying dangerous behaviors.
Because we have access to a simulator of the robot, one could think that the safety of a potential behavior could be evaluated using the simulation; however, this is likely to not be sufficient because high-performing controllers for the intact robot might have considerably different behavior and often be harmful on the damaged robot. More often than not, such behaviors are highly dependent on the robot's legs, wheels, etc., the failure of which would result in a radically different outcome in the damaged case. IT\&E having no prior knowledge about the damage or any safety limitations, will most likely attempt more than a few of these dangerous behaviors. These behaviors constitute a big risk for further damaging fragile robots. To make matters worse, a damage recovery algorithm would have to compensate for the reality gap~\citep{koos2013transferability, jakobi1995noise} as well\footnote{The reality gap refers to the differences between simulated and physical systems.}. In particular, even for intact robots, the behaviors contained in the behavior-performance maps are not likely to be reproduced exactly on the physical robot.

We address these issues by introducing a safety-aware IT\&E algorithm (sIT\&E; see Fig.~\ref{fig:concept}). sIT\&E uses a constrained BO procedure~\citep{gardner2014bayesian} in which each user-defined safety constraint, $c_i(\mathbf{x})$, is modeled as a separate GP. The next sample is selected by optimizing the Expected Constrained Improvement (ECI) acquisition function~\citep{gardner2014bayesian}:
\begin{equation}
    ECI(\mathbf{\hat{x}}) = EI(\mathbf{\hat{x}})\prod_{i=1}^n p(c_i(\mathbf{\hat{x}}) \geq 0)
\end{equation}
where $\mathbf{\hat{x}}$ is the candidate point, $c_i(\mathbf{\hat{x}}), i \in \{ 1, \cdots , n \}$ are the $n$ constraint functions and $EI(\mathbf{\hat{x}})$ is the standard expected improvement~\citep{gardner2014bayesian}.

The differences between sIT\&E (see Alg.\ref{algo:site}) and IT\&E are as follows: (1) the behavior descriptor in MAP-Elites is augmented with extra dimensions for each safety constraint, so that sIT\&E will have a good estimate of the safe regions (i.e., where all inequality safety constraints are fulfilled); for example, these dimensions could be torque or IMU measurements that should not exceed a specific threshold; (2) during the on-line adaptation step, sIT\&E optimizes for performance while also guiding the search through the safe regions.
%

\begin{algorithm}
\caption{Safety-aware Intelligent Trial-and-Error (sIT\&E)}\label{euclid}
\label{algo:site}
\begin{algorithmic}
    \Procedure{sIT\&E}{}
        \State \text{Before mission (intact robot in simulation):}
        \State \quad \text{Create Behavior-Performance Map via MAP-Elites storing safety related information}
        \While{in mission}
            \If{significant performance drop}
                \State Adaptation Step (via M-CBO Algorithm)
            \EndIf
        \EndWhile
    \EndProcedure

    \Procedure{Map-Based Constrained Bayesian Optimization (M-CBO)}{}
        \State \text{$\forall \mathbf{x} \in map:$}
        \State \quad $p(f(\mathbf{x}) | \mathbf{x}) = N(m(\mathbf{x}), k(\mathbf{x}, \mathbf{x}))$
        \State \quad $p(c_i(\mathbf{x}) | \mathbf{x}) = N(m_{c_i}(\mathbf{x}), k_{c_i}(\mathbf{x}, \mathbf{x}))$, $i \in \{ 1, \cdots , n \}$

        \While{stopping criteria not met}
            \State $\mathbf{x}_{t+1} = \argmax_{\mathbf{x}}ECI(\mathbf{x}|X_{1:t},C_{1:t})$
            \State $\{c_1(\mathbf{x}_{t+1}),~\dots~, c_n(\mathbf{x}_{t+1}), f(\mathbf{x}_{t+1})\} = execute\_behavior(\mathbf{x}_{t+1})$
            \State $X_{1:t+1} = \{ f(\mathbf{x}_{t+1}), X_{1:t} \}$
            \State $C_{1:t+1} = \{ \langle c_1(\mathbf{x}_{t+1}),~\dots~, c_n(\mathbf{x}_{t+1})\rangle, C_{1:t} \}$
            \State \text{Update GPs for the objective function/safety constraints}
        \EndWhile
    \EndProcedure
\end{algorithmic}
\end{algorithm}

\section{Crawling humanoid robot experiments}

To evaluate our algorithm, we use a simulated iCub robot~\citep{metta2008icub, tsagarakis2007icub} performing a crawling task.
Learning how to crawl could prove especially useful in humanoid robot damage recovery, where attempting to walk again might constitute a big risk for further damages or be infeasible altogether (e.g. traversing a short or tight tunnel). Furthermore, solving this task serves as a stepping stone towards damage recovery for more advanced tasks (e.g. walking).
%
%

To generate a diversity of behaviors with MAP-Elites, we augment an initially 4D behavior descriptor, defined as the fraction of time each arm/leg spent on the ground, with a safety dimension that encodes the sum of contact point forces.
%
%
%
sIT\&E optimizes for crawling speed and is constrained by a safety threshold for the sum of contact point forces. This threshold is determined after conducting several preliminary experiments in order to understand the correlation between the robot's behavior and the contact point forces at high crawling speeds.
%
%
To optimize the acquisition function, we iterate over all the points in the behavior-performance map (which contains approximately 1500 behaviors), and select a behavior that is estimated to be the most promising above the safety threshold.

We compare 3 algorithms in terms of the best safe performance observed and unsafe trials attempted: (1) IT\&E maximizing crawling speed; (2) a multi-objective~\citep{deb2014multi} IT\&E algorithm (MO-IT\&E; based on the Expected Hypervolume Improvement~\citep{yang2015expected, hupkens2015faster}), that maximizes the crawling speed and minimizes the sum of contact point forces, therefore, building a Pareto front from which the safest behavior can be chosen; and (3) sIT\&E maximizing crawling speed within the safe region as described above. We test 4 damage conditions: (1) locked shoulder joint, (2) locked hip joint, (3) locked shoulder joint \& angled elbow, and (4) combination of 2 \& 3.

\begin{figure}[h]
    \centering
    \includegraphics[width=0.95\textwidth]{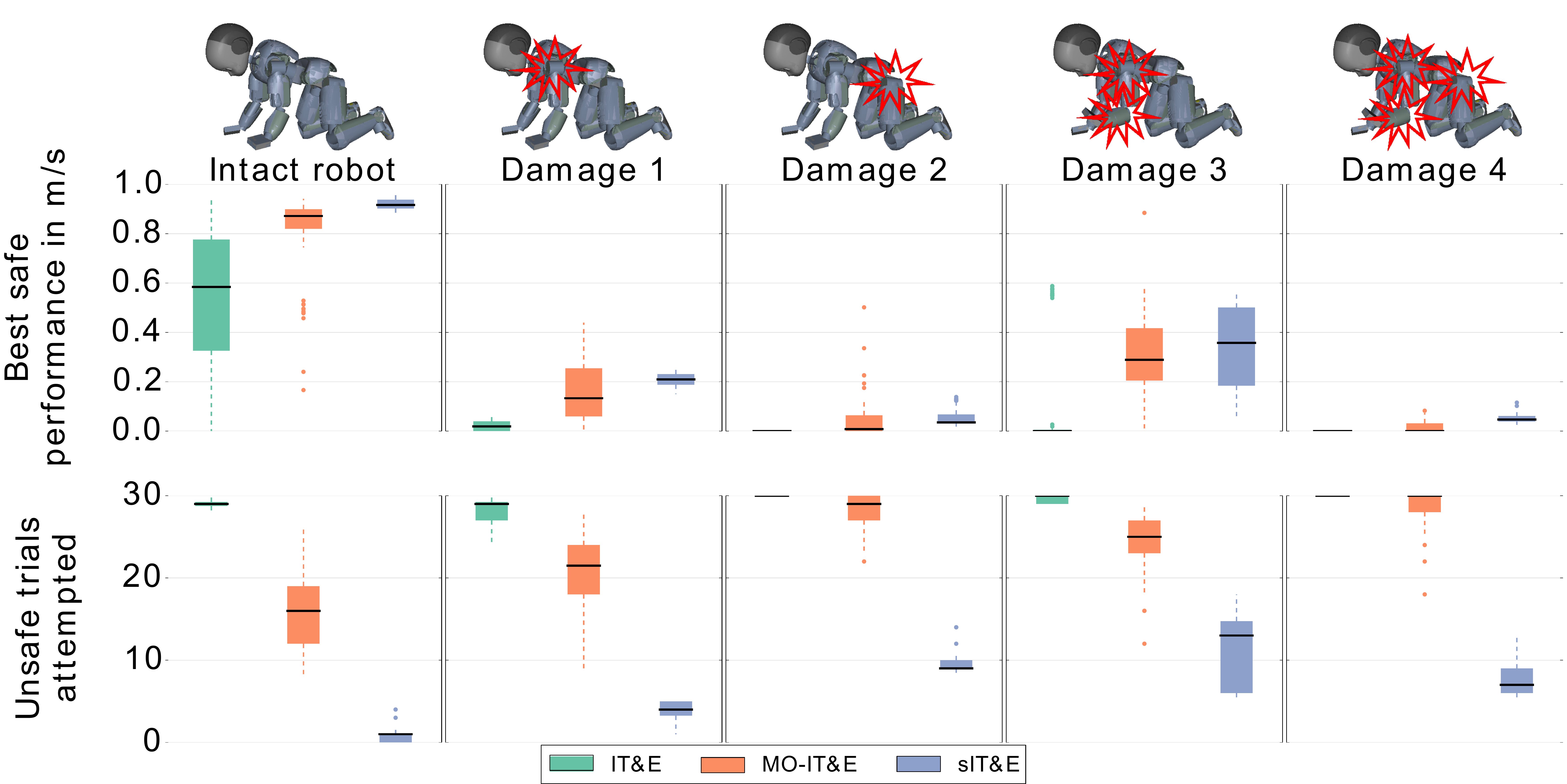}
    \caption{\textbf{Comparison between IT\&E, MO-IT\&E and sIT\&E}. Each algorithm is ran 20 times for 4 different behavior-performance maps. We report the best safe performance (i.e., the fastest crawling speed within the safe region observed during the learning process) (upper row) and number of unsafe trials attempted (violating the constraints) (lower row). Horizontal black lines represent medians.}
    \label{fig:results}
\end{figure}
To avoid depending on a single behavior-performance map, we use 4 independently-generated maps and run each algorithm 20 times for 30 trials. We use the squared exponential kernel with $\sigma = 0.1$ and a GP noise value of 0.01.
%
%
%
When using IT\&E, the median number of dangerous trials is approximately equal to 29 (out of 30) in all damage settings (Fig.~\ref{fig:results}, lower row). For MO-IT\&E, this number decreases, but it is still greater than 22. In contrast, sIT\&E requires less than 10 unsafe trials for damages 1, 2, and 4, and 14 for damage 3. Pairwise comparisons indicate that the results are highly significant ($p<0.0001$, Mann-Whitney U test).
In terms of safe performance (crawling speed in m/s), sIT\&E dominates over both IT\&E and MO-IT\&E in all damage settings (Fig.~\ref{fig:results}, upper row), with the results being statistically significant ($p<0.001$) in all cases apart from damage 3.
%
%
%

All experiments were conducted using the limbo framework~\citep{cully2016limbo}. A supplementary video is available at \url{https://youtu.be/8esrj-7WhsQ}.

\section{Conclusion}
Our experiments show that the vanilla IT\&E algorithm finds high-performing behaviors in a few trials, but most of the behaviors tested, including the best, final ones, are unsafe for the robot. Since the multi-objective approach searches for a set of Pareto-optimal trade-offs, it can find safe and high-performing behaviors; however, this approach still tests many unsafe behaviors during the learning phase. By contrast, the sIT\&E algorithm finds gaits that are both safe and high-performing with only a handful of unsafe trials. Thanks to this property, we are confident that sIT\&E is less likely to damage the real iCub than IT\&E or BO. In future work, we will compare our results to approaches based on safety regions~\citep{berkenkamp2016bayesian}, which might be safer, but may prove too conservative performance-wise. Overall, this work shows that safety is a critical component for any robot learning algorithm and that constrained BO can provide a good basis to design algorithms that are both data-efficient and safe.

\subsubsection*{Acknowledgments}

This work received funding from the European Research Council (ERC) under the European Union’s Horizon 2020 research and innovation program (grant agreement number 637972, project \enquote{ResiBots}).

\small
\bibliographystyle{plain}
\bibliography{nips_2016}

\end{document}